\documentclass[11pt,a4paper]{article}
\usepackage[hyperref]{acl2018}
\usepackage{times}
\usepackage{latexsym}
\usepackage{multirow} 
\usepackage{booktabs} 

\usepackage{url}

\usepackage{graphics}
\usepackage{graphicx}
\usepackage{subcaption}
\usepackage{makecell}  
\usepackage{footnote}  
\makesavenoteenv{tabular}
\makesavenoteenv{table}

\aclfinalcopy 

\title{Reference-less Quality Estimation of Text Simplification Systems}

\newcommand{\email}[1]{#1}
 
\author{Louis Martin \\
  Facebook AI Research \& Inria \\
  \email{louismartin@fb.com} \\\And
  Samuel Humeau \\
  Facebook AI Research  \\
  \email{samuelhumeau@fb.com} \\\And
  Pierre-Emmanuel Mazar\'e \\
  Facebook AI Research  \\
  \email{pem@fb.com} \\\AND
  Antoine Bordes \\
  Facebook AI Research  \\
  \email{abordes@fb.com} \\\And
  \'Eric de La Clergerie\\
  Inria  \\
  \email{eric.de\_la\_clergerie@inria.fr}\\\And
  Beno{\^\i}t Sagot\\
  Inria\\
  \email{benoit.sagot@inria.fr}\\
  }

\date{}

\begin{document}
\maketitle
\begin{abstract}
    The evaluation of text simplification (TS) systems remains an open challenge. As the task has common points with machine translation (MT), TS is often evaluated using MT metrics such as BLEU. However, such metrics require high quality reference data, which is rarely available for TS. TS has the advantage over MT of being a monolingual task, which allows for direct comparisons to be made between the simplified text and its original version. In this paper, we compare multiple approaches to reference-less quality estimation of sentence-level text simplification systems, based on the dataset used for the QATS 2016 shared task. We distinguish three different dimensions: grammaticality, meaning preservation and simplicity. We show that $n$-gram-based MT metrics such as BLEU and METEOR correlate the most with human judgment of grammaticality and meaning preservation, whereas simplicity is best evaluated by basic length-based metrics.
\end{abstract}


\section{Introduction}
Text simplification (hereafter TS) has received increasing interest by the scientific community in recent years.
It aims at producing a simpler version of a source text that is both easier to read and to understand, thus improving the accessibility of text for people suffering from a range of disabilities such as aphasia \cite{carroll1998practical} or dyslexia \cite{rello2013simplify}, as well as for second language learners \cite{xia2016text} and people with low literacy \cite{watanabe2009facilita}.
This topic has been researched for a variety of languages such as English \cite{zhu2010monolingual, wubben2012sentence, narayan2014hybrid, xu2015problems}, French \cite{brouwers2014syntactic}, Spanish \cite{saggion2011text}, Portuguese \cite{specia2010translating}, Italian \cite{brunato2015design} and Japanese \cite{goto2015japanese}.\footnote{Note that text simplification has also been used as a pre-processing step for other natural language processing tasks such as machine translation \cite{chandrasekar1996motivations} and semantic role labelling \cite{vickrey2008sentence}.}

One of the main challenges in TS is finding an adequate automatic evaluation metric, which is necessary to avoid the time-consuming human evaluation. Any TS evaluation metric should take into account three properties expected from the output of a TS system, namely:
\begin{itemize}
    \item Grammaticality: how grammatically correct is the TS system output?
    \item Meaning preservation: how well is the meaning of the source sentence preserved in the TS system output?
    \item Simplicity: how simple is the TS system output?\footnote{There is no unique way to define the notion of {\em simplicity} in this context. Previous works often rely on the intuition of human annotators to evaluate the level of simplicity of a TS system output.}
\end{itemize}

TS is often reduced to a sentence-level problem, whereby one sentence is transformed into a simpler version containing one or more sentences. In this paper, we shall make use of the terms {\em source (sentence)} and {\em (TS system) output} to respectively denote a sentence given as an input to a TS system and the simplified, single or multi-sentence output produced by the system.

TS, seen as a sentence-level problem, is often viewed as a monolingual variant of  (sentence-level) MT. The standard approach to automatic TS evaluation is therefore to view the task as a translation problem and to use machine translation (MT) evaluation metrics such as BLEU \cite{papineni2002bleu}.
However, MT evaluation metrics rely on the existence of parallel corpora of source sentences and manually produced reference translations, which are available on a large scale for many language pairs \cite{tiedemann12}. 
TS datasets are less numerous and smaller.
Moreover, they are often automatically extracted from comparable corpora rather than strictly parallel corpora, which results in noisier reference data.
For example, the PWKP dataset \cite{zhu2010monolingual} consists of 100,000 sentences from the English Wikipedia automatically aligned with sentences from the Simple English Wikipedia based on term-based similarity metrics.
It has been shown by \citet{xu2015problems} that many of PWKP's ``simplified'' sentences are in fact not simpler or even not related to their corresponding source sentence.
Even if better quality corpora such as Newsela do exist \cite{xu2015problems}, they are costly to create, often of limited size, and not necessarily open-access.

This creates a challenge for the use of reference-based MT metrics for TS evaluation.
However, TS has the advantage of being a monolingual translation-like task, the source being in the same language as the output. This allows for new, non-conventional ways to use MT evaluation metrics, namely by using them to compare the output of a TS system with the source sentence, thus avoiding the need for reference data. However, such an evaluation method can only capture at most two of the three above-mentioned dimensions, namely meaning preservation and, to a lesser extent, grammaticality.


Previous works on reference-less TS evaluation include \citet{vstajner2014one}, who compare the behaviour of six different MT metrics when used between the source sentence and the corresponding simplified output. They evaluate these metrics with respect to meaning preservation and grammaticality.
We extend their work in two directions. Firstly, we extend the comparison to include the degree of simplicity achieved by the system. Secondly, we compare additional features, including those used by \citet{vstajner2016quality}, both individually, as elementary metrics, and within multi-feature metrics.
To our knowledge, no previous work has provided as thorough a comparison across such a wide range and combination of features for the reference-less evaluation of TS.


    
First we review available text simplification evaluation methods and traditional quality estimation features.
We then present the QATS shared task and the associated dataset, which we use for our experiments.
Finally we compare all methods in a reference-less setting and analyze the results.

\section{Existing evaluation methods}

\subsection{Using MT metrics to compare the output and a reference}
TS can be considered as a monolingual translation task. As a result, MT metrics such as BLEU \cite{papineni2002bleu}, which compare the output of an MT system to a reference translation, have been extensively used for TS \cite{narayan2014hybrid, vstajner2015deeper, xu2016optimizing}.
Other successful MT metrics include TER \cite{snover2009ter}, ROUGE \cite{lin2004rouge} and METEOR \cite{banerjee2005meteor}, but they have not gained much traction in the TS literature.

These metrics rely on good quality references, something which is often not available in TS, as discussed by \citet{xu2015problems}.
Moreover, \citet{vstajner2015deeper} and \citet{sulem2018bleu} showed that using BLEU to compare the system output with a reference is not a good way to perform TS evaluation, even when good quality references are available.
This is especially true when the TS system produces more than one sentence for a single source sentence.

\subsection{Using MT metrics to compare the output and the source sentence}
As mentioned in the Introduction, the fact that TS is a monolingual task means that MT metrics can also be used to compare a system output with its corresponding source sentence, thus avoiding the need for reference data.
Following this idea, \citet{vstajner2014one} found encouraging correlations between 6 widely used MT metrics and human assessments of grammaticality and meaning preservation.
However MT metrics are not relevant for the evaluation of simplicity, which is why they did not take this dimension into account.
\citet{xu2016optimizing} also explored the idea of comparing the TS system output with its corresponding source sentence, but their metric, SARI, also requires to compare the output with a reference. In fact, this metric is designed to take advantage of more than one reference. It can be applied when only one reference is available for each source sentence, but its results are better when multiple references are available.

Attempts to perform Quality Estimation on the output of TS systems, without using references, include the 2016 Quality Assessment for Text Simplification (QATS) shared task \cite{vstajner2016shared}, to which we shall come back in section \ref{methodology}.
\citet{sulem2018semantic} introduce another approach, named SAMSA. The idea is to evaluate the structural simplicity of a TS system output given the corresponding source sentence.
SAMSA is maximized when the simplified text is a sequence of short and simple sentences, each accounting for one semantic event in the original sentence. It relies on an in-depth analysis of the source sentence and the corresponding output, based on a semantic parser and a word aligner. A drawback of this approach is that good quality semantic parsers are only available for a handful of languages.
The intuition that sentence splitting is an important sub-task for producing simplified text motivated \citet{narayan2017split} to organize the  \textit{Split and Rephrase} shared task, which was dedicated to this problem.

\subsection{Other metrics}
One can also estimate the quality of a TS system output based on simple features extracted from it.

For instance, the \textsc{QuEst} framework for quality estimation in MT gives a number of useful baseline features for evaluating an output sentence \cite{specia2013quest}.
These features range from simple statistics, such as the number of words in the sentence, to more sophisticated features, such as the probability of the sentence according to a language model.
Several teams who participated in the QATS shared task used metrics based on this framework, namely SMH \cite{vstajner2016quality}, UoLGP \cite{rios2015large} and UoW \cite{bechara2015miniexperts}.

Readability metrics such as Flesch-Kincaid Grade Level (FKGL) and Flesch Reading Ease (FRE) \cite{kincaid1975derivation} have been extensively used for evaluating simplicity.
These two metrics, which were shown experimentally to give good results, are linear combinations of the number of words per sentence and the number of syllables per word, using carefully adjusted weights.
 

\section{Methodology} \label{methodology}
Our goal is to compare a large number of ways to perform TS evaluation without a reference. To this end, we use the dataset provided in the QATS shared task.
We first compare the behaviour of elementary metrics, which range from commonly used metrics such as BLEU to basic metrics based on a single low-level feature such as sentence length.
We then compare the effect of aggregating these elementary metrics into more complex ones and compare our results with the state of the art, based on the QATS shared task data and results.

\subsection{The QATS shared task}
\begin{figure}
    \centering
    \includegraphics[width=\columnwidth]{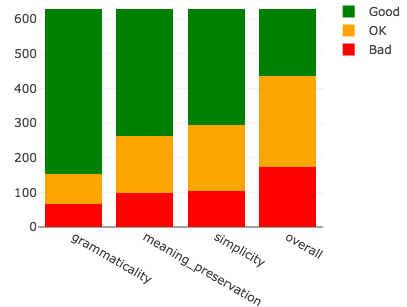}
    \caption{Label repartition on the QATS Shared task}
    \label{label_repartition}
\end{figure}

The data from the QATS shared task \cite{vstajner2016shared} consists of a collection of 631 pairs of english sentences composed of a source sentence extracted from an online corpus and a simplified version thereof, which can contain one or more sentences.
This collection is split into a training set (505 sentence pairs) and a test set (126 sentence pairs).
Simplified versions were produced automatically using one of several TS systems trained by the shared task organizers.
Human annotators labelled each sentence pair using one of the three labels {\em Good}, {\em OK} and {\em Bad} on each of the three dimensions: grammaticality, meaning preservation and simplicity\footnote{We were not able to find detailed information about the annotation process. In particular, we do not know whether each sentence was annotated only once or whether multiple annotations were produced, followed by an adjudication step.}.
An overall quality label was then automatically assigned to each sentence pair based on its three manually assigned labels using a method detailed in \cite{vstajner2016shared}.
Distribution of the labels and examples are presented in FIGURE~\ref{label_repartition} and TABLE~\ref{qats_examples}.

\renewcommand\cellgape{\Gape[2pt]}
\begin{table*}
\small
\centering
\scalebox{0.95}{
\begin{tabular}{@{}llccccl@{}}
\toprule
\multirow{2}{*}{Version} & \multirow{2}{*}{Sentence} & \multicolumn{4}{c}{Aspect} & \multirow{2}{*}{Modification} \\
&& G & M & S & O & \\
\midrule
Original & \makecell[l]{All three were arrested in the Toome area \textbf{and} have been taken \\ to the Serious Crime Suite at Antrim police station.} & \multirow{2}{*}{good} & \multirow{2}{*}{good} & \multirow{2}{*}{good} & \multirow{2}{*}{good} & \multirow{2}{*}{syntactic} \\
Simple & \makecell[l]{All three were arrested in the Toome area. \textbf{All three} have been \\ taken to the Serious Crime Suite at Antrim police station.} \\
\midrule
Original & \makecell[l]{For years the former Bosnia Serb army commander Ratko \\ Mladic had evaded capture \textbf{and was one of the world’s most} \\ \textbf{wanted men, but his time on the run finally ended last year} \\ \textbf{when he was arrested near Belgrade.}} & \multirow{2}{*}{good} & \multirow{2}{*}{bad} & \multirow{2}{*}{ok} & \multirow{2}{*}{bad} & \multirow{2}{*}{content reduction} \\
Simple & \makecell[l]{For years the former Bosnia Serb army commander Ratko \\ Mladic had evaded capture.} \\
\midrule
Original & \makecell[l]{Madrid was occupied by French troops during the Napoleonic \\ Wars, and Napoleon’s brother Joseph was \textbf{installed} on the \\ throne.} & \multirow{2}{*}{good} & \multirow{2}{*}{good} & \multirow{2}{*}{good} & \multirow{2}{*}{good} & \multirow{2}{*}{lexical} \\
Simple & \makecell[l]{Madrid was occupied by French troops during the Napoleonic \\ Wars, and Napoleon’s brother Joseph was \textbf{put} on the throne.} \\
\midrule
Original & \makecell[l]{Keeping articles with potential \textbf{encourages} editors, especially \\ unregistered users, to be bold and improve the article to allow it \\ to evolve over time.} & \multirow{2}{*}{bad} & \multirow{2}{*}{bad} & \multirow{2}{*}{ok} & \multirow{2}{*}{bad} & \multirow{2}{*}{dropping} \\
Simple & \makecell[l]{Keeping articles with potential editors, especially unregistered \\ users, to be bold and improve the article to allow it to evolve \\ over time.} \\
\bottomrule
\end{tabular}
}
\caption{Examples from the training dataset of QATS. Differences between the original and the simplified version are presented in bold. This table is adapted from \citet{vstajner2016shared}.}
\label{qats_examples}
\end{table*}

The goal of the shared task is, for each sentence in the test set, to either produce a label ({\em Good}, {\em OK}, {\em Bad}) or a raw score estimating the overall quality of the simplification for each of the three dimensions.
Raw score predictions are evaluated using the Pearson correlation with the ground truth labels, while actual label prediction are evaluated using the weighted F1-score.
The shared task is described in further details on the QATS website\footnote{\url{http://qats2016.github.io/shared.html}}.

\subsection{Features}
\label{feature_list}
In our experiments, we compared about 60 elementary metrics, which can be organised as follows:
\begin{itemize}
    \item MT metrics
    \begin{itemize}
        \item BLEU, ROUGE, METEOR, TERp
        \item Variants of BLEU: BLEU\_1gram, BLEU\_2gram, BLEU\_3gram, BLEU\_4gram and seven smoothing methods\footnote{\url{https://www.nltk.org/api/nltk.translate.html\#nltk.translate.bleu\_score.SmoothingFunction}} from NLTK \cite{bird2004nltk}.
        \item Intermediate components of TERp inspired by \cite{vstajner2016quality}: e.g. number of insertions, deletions, shifts...
    \end{itemize}
    \item Readability metrics and other sentence-level features: FKGL and FRE, numbers of words, characters, syllables...
    \item Metrics based on the baseline \textsc{QuEst} features (17 features) \cite{specia2013quest}, such as statistics on the number of words, word lengths, language model probability and $n$-gram frequency.
    \item Metrics based on other features: frequency table position, concreteness as extracted from \citeauthor{brysbaert2014concreteness}'s \citeyear{brysbaert2014concreteness} list, language model probability of words using a convolutional sequence to sequence model from \cite{gehring2017convolutional}, comparison methods using pre-trained fastText word embeddings \cite{mikolov2018advances} or Skip-thought sentence embeddings \cite{kiros2015skip}.
\end{itemize}
TABLE~\ref{features} lists 30 of the elementary metrics that we compared, which are those that we found to correlate the most with human judgments on one or more of the three dimensions (grammaticality, meaning preservation, simplicity).

\subsection{Experimental setup}
\paragraph{Evaluation of elementary metrics}
We rank all features by comparing their behaviour with human judgments on the training set.
We first compute for each elementary metric the Pearson correlation between its results and the manually assigned labels for each of the three dimensions.
We then rank our elementary  metrics according to the absolute value of the Pearson correlation.\footnote{The code is available on Github at \url{https://github.com/facebookresearch/text-simplification-evaluation}}

\paragraph{Training and evaluation of a combined metric}
We use our elementary metrics as features to train classifiers on the training set, and evaluate their performance on the test set.
We therefore scale them and reduce the dimensionality with a 25-component PCA\footnote{We used PCA instead of feature selection because it performed better on the validation set. The number of component was tuned on the validation set as well.}, then train several regression algorithms\footnote{Regressors: Linear regression, Lasso, Ridge, Linear SVR (SVM regressor), Adaboost regressor, Gradient boosting regressor and Random forest regressor.} and classification algorithms\footnote{Classifiers: Logistic regression, MLP classifier (with L2 penalty, alpha=1), SVC (linear SVM classifier), K-nearsest neighbors classifier (k=3), Adaboost classifier, Gradient boosting classifier and Random forest classifier.} using scikit-learn \cite{pedregosa2011}.
For each dimension, we keep the two models performing best on the test set and add them in the leaderboard of the QATS shared task (TABLE~\ref{leaderboard}), naming them with the name of the regression algorithm they were built with.

\begin{table*}
\scalebox{0.95}{
\begin{tabular}{ll}
\toprule
               Short name &                                                                   Description \\
\midrule
 NBSourcePunct &  Number of punctuation tokens in source (\textsc{QuEst}) \\
 NBSourceWords &  Number of source words (\textsc{QuEst}) \\
 NBOutputPunct &  Number of punctuation tokens in output (\textsc{QuEst}) \\
 TypeTokenRatio &  Type token ratio (\textsc{QuEst}) \\
 TERp\_Del &  Number of deletions (TERp component) \\
 TERp\_NumEr &  Number of total errors (TERp component) \\
 TERp\_Sub &  Number of substitutions (TERp component) \\
 TERp &  TERp MT metric \\
 BLEU\_1gram &  BLEU MT metric with unigrams only \\
 BLEU\_2gram &  BLEU MT metric up to bigrams \\
 BLEU\_3gram &  BLEU MT metric up to trigrams \\
 BLEU\_4gram &  BLEU MT metric up to 4-grams \\
 METEOR &  METEOR MT metric \\
 ROUGE &  ROUGE summarization metric \\
 BLEUSmoothed &  BLEU MT metric with smoothing (method 7 from nltk) \\
 AvgCosineSim &  Cosine similarity between source and output pre-trained word embeddings\\
 NBOutputChars &  Number of characters in the output \\
 NBOutputCharsPerSent &  Average number of characters per sentence in the output \\
 NBOutputSyllables &  Number of syllables in the output \\
 NBOutputSyllablesPerSent &  Average number of syllables per sentence in the output \\
 NBOutputWords &  Number of words in the output \\
 NBOutputWordsPerSent &  Average number of words per sentence in the output \\
 AvgLMProbsOutput &  Average log-probabilities of output words (Language Model) \\
 MinLMProbsOutput &  Minimum log-probability of output words (Language Model) \\
 MaxPosInFreqTable &  Maximum position of output words in the frequency table \\
 AvgConcreteness &  Average word concreteness \citeauthor{brysbaert2014concreteness}'s \citeyear{brysbaert2014concreteness} concreteness list \\
 OutputFKGL &  Flesch-Kincaid Grade Level \\
 OutputFRE &  Flesch Reading Ease \\
     WordsInCommon &  Percentage of words in common between source and Output \\
\bottomrule
\end{tabular}
}
\caption{Brief description of 30 of our most relevant elementary metrics}
\label{features}
\end{table*}

\section{Results}

\subsection{Comparing elementary metrics}

\begin{table*}
\small
\centering
\scalebox{0.95}{
\begin{tabular}{@{}lrrlrrlrr@{}}
\toprule
\multicolumn{3}{@{}l}{Grammaticality} & \multicolumn{3}{l}{Meaning Preservation} & \multicolumn{3}{l}{Simplicity} \\
    Short name & Train $\downarrow$ &  Test &            Short name & Train $\downarrow$ &  Test &            Short name & Train $\downarrow$ &  Test \\
\midrule
 \textit{Best QATS team} & & 0.48 & \textit{Best QATS team} & & 0.59 & \textit{Best QATS team} & & 0.38 \\
 METEOR & 0.36 & 0.39 &  BLEUSmoothed & 0.59 & 0.52 &  NBOutputCharsPerSent & -0.52 & -0.45 \\
 BLEUSmoothed & 0.33 & 0.34 &  BLEU\_3gram & 0.57 & 0.52 &  NBOutputSyllablesPerSent & -0.52 & -0.49 \\
 BLEU\_4gram & 0.32 & 0.34 &  METEOR & 0.57 & 0.58 &  NBOutputWordsPerSent & -0.51 & -0.39 \\
 BLEU\_3gram & 0.31 & 0.34 &  BLEU\_2gram & 0.57 & 0.52 &  NBOutputChars & -0.48 & -0.37 \\
 TERp\_NumEr & -0.30 & -0.31 &  BLEU\_4gram & 0.57 & 0.51 &  NBOutputWords & -0.47 & -0.29 \\
 BLEU\_2gram & 0.30 & 0.34 &  WordsInCommon & 0.55 & 0.50 &  NBOutputSyllables & -0.46 & -0.42 \\
 TERp & -0.30 & -0.32 &  BLEU\_1gram & 0.55 & 0.52 &  NBOutputPunt & -0.42 & -0.31 \\
 ROUGE & 0.29 & 0.29 &  ROUGE & 0.55 & 0.47 &  NBSourceWords & -0.38 & -0.21 \\
 AvgLMProbsOutput & 0.28 & 0.34 &  TERp & -0.54 & -0.48 &  outputFKGL & -0.36 & -0.37 \\
 BLEU\_1gram & 0.27 & 0.33 &  TERp\_NumEr & -0.53 & -0.49 &  NBSourcePunct & -0.34 & -0.18 \\
 WordsInCommon & 0.27 & 0.30 &  TERp\_Del & -0.50 & -0.52 &  TypeTokenRatio & -0.22 & -0.04 \\
 TERp\_Del & -0.27 & -0.35 &  AvgCosineSim & 0.44 & 0.34 &  AvgConcreteness & 0.21 & 0.32 \\
 NBSourceWords & -0.25 & -0.07 &  AvgLMProbsOutput & 0.39 & 0.36 &  MaxPosInFreqTable & -0.18 & 0.03 \\
 AvgCosineSim & 0.23 & 0.25 &  AvgConcreteness & -0.28 & -0.06 &  MinLMProbsOutput & 0.17 & 0.15 \\
 MinLMProbsOutput & 0.11 & -0.07 &  NBSourceWords & -0.28 & -0.13 &  OutputFRE & 0.16 & 0.27 \\ 
 \bottomrule
\end{tabular}
}
\caption{Pearson correlation with human judgments of elementary metrics ranked by absolute value on training set (15 best metrics for each dimension).}
\label{individual_features}
\end{table*}

FIGURE~\ref{individual_features} ranks all elementary metrics given their absolute Pearson correlation on each of the three dimensions.

\paragraph{Grammaticality}

$N$-gram based MT metrics have the highest correlation with human grammaticality judgments.
METEOR seems to be the best, probably because of its robustness to synonymy, followed by smoothed BLEU (BLEUSmoothed in~\ref{features}).
This indicates that relevant grammaticality information can be derived from the source sentence.
We were expecting that information contained in a language model would help achieving better results (\textit{AvgLMProbsOutput}), but MT metrics correlate better with human judgments.
We deduce that the grammaticality information contained in the source is more specific and more helpful for evaluation than what is learned by the language model.

\paragraph{Meaning preservation}
It is not surprising that meaning preservation is best evaluated using MT metrics that compare the source sentence to the output sentence, with in particular smoothed BLEU, BLEU\_3gram and METEOR.
Very simple features such as the percentage of words in common between source and output also rank high.
Surprisingly, word embedding comparison methods do not perform as well for meaning preservation, even when using word alignment.

\paragraph{Simplicity}
Methods that give the best results are the most straightforward for assessing simplicity, namely word, character and syllable counts in the output, averaged over the number of output sentences.
These simple features even outperform the traditional, more complex metrics FKGL and FRE.
As could be expected, we find that metrics with the highest correlation to human simplicity judgments only take the output into account.
Exceptions are the \textit{NBSourceWords} and \textit{NBSourcePunct} features.
Indeed, if the source sentence has a lot of words and punctuation, and is therefore likely to be particularly complex, then the output will most likely be less simple as well.
We also expected word concreteness ratings and position in the frequency table to be good indicators of simplicity, but it does not seem to be the case here.
Structural simplicity might simply be more important than such more sophisticated components of the human intuition of simple text.

\paragraph{Discussion}
Even if counting the number of words or comparing $n$-grams are good proxies for the simplification quality, they are still very superficial features and might miss some deeper and more complex information.
Moreover the fact that grammaticality and meaning preservation are best evaluated using $n$-gram-based comparison metrics might bias the TS models towards copying the source sentence and applying fewer modifications.

Syntactic parsing or language modelling might capture more insightful grammatical information and allow for more flexibility in the simplification model.
Regarding meaning preservation, semantic analysis or paraphrase detection models would also be good candidates for a deeper analysis.

\paragraph{Warning note}

We should be careful when interpreting these results as the QATS dataset is relatively small.
We compute confidence intervals on our results, and find them to be non-negligible, yet without putting our general observations into question. 
For instance, \mbox{METEOR}, which performs best on grammaticality, has a $95\%$ confidence interval of $0.36 \pm 0.08$ on the training set.
These results are therefore preliminary and should be validated on other datasets.

\subsection{Combination of all features with trained models}
We also combine all elementary metrics and train an evaluation models for each of the three dimensions. TABLE~\ref{leaderboard_pearson} presents our two best regressors in validation for each of the dimensions and TABLE~\ref{leaderboard_f1} for classifiers.

\begin{table*}
\begin{subtable}{\textwidth}
\centering\small
\begin{tabular}{llll}
\toprule
                     Grammaticality &
                     Meaning Preservation &                          Simplicity &
                     Overall \\
\midrule
 0.482   OSVCML1 &  0.588   IIT-Meteor &  0.487   \textbf{Ridge} &  0.423   \textbf{Ridge} \\
 0.384   METEOR &  0.585   OSVCML &  0.456   \textbf{LinearSVR} &  0.423   \textbf{LinearRegression} \\
 0.344   BLEU &  0.575   \textbf{Ridge} &  0.382   OSVCML1 &  0.343   OSVCML2 \\
 0.340   OSVCML &  0.573   OSVCML2 &  0.376   OSVCML2 &  0.334   OSVCML \\
 0.327   \textbf{Lasso} &  0.555   \textbf{Lasso} &  0.339   OSVCML &  0.232   SimpleNets-RNN2 \\
 0.323   TER &  0.533   BLEU &  0.320   SimpleNets-MLP &  0.230   OSVCML1 \\
 0.308   SimpleNets-MLP &  0.527   METEOR &  0.307   SimpleNets-RNN3 &  0.205   UoLGP-emb \\
 0.308   WER &  0.513   TER &  0.240   SimpleNets-RNN2 &  0.198   SimpleNets-MLP \\
 0.256   UoLGP-emb &  0.495   WER &  0.123   UoLGP-combo &  0.196   METEOR \\
 0.256   UoLGP-combo &  0.482   OSVCML1 &  0.120   UoLGP-emb &  0.189   UoLGP-combo \\
 0.208   UoLGP-quest &  0.465   SimpleNets-MLP &  0.086   UoLGP-quest &  0.144   UoLGP-quest \\
 0.118   \textbf{GradientBoostingRegressor} &  0.285   UoLGP-quest &  0.052   IIT-S &  0.130   TER \\
 0.064   SimpleNets-RNN3 &  0.262   SimpleNets-RNN2 &  -0.169   METEOR &  0.112   SimpleNets-RNN3 \\
 0.056   SimpleNets-RNN2 &  0.262   SimpleNets-RNN3 &  -0.242   TER &  0.111   WER \\
   &  0.250   UoLGP-combo &  -0.260   WER &  0.107   BLEU \\
   &  0.188   UoLGP-emb &  -0.267   BLEU &    \\
\bottomrule
\end{tabular}
\caption{Pearson correlation for regressors (raw scoring)}
\label{leaderboard_pearson}
\end{subtable}

\vspace{2ex}

\begin{subtable}{\textwidth}
\centering\small
\begin{tabular}{llll}
\toprule
                     Grammaticality &
                     Meaning Preservation &                          Simplicity &
                     Overall \\
\midrule
 71.84   SMH-RandForest &  70.14   \textbf{SVC} &  61.60   \textbf{SVC} &  49.61   \textbf{LogisticRegression} \\
 71.64   SMH-IBk &  68.07   SMH-Logistic &  56.95   \textbf{AdaBoostClassifier} &  48.57   SMH-RandForest-b \\
 70.43   \textbf{LogisticRegression} &  65.60   MS-RandForest &  56.42   SMH-RandForest-b &  48.20   UoW \\
 69.96   SMH-RandForest-b &  64.40   SMH-RandForest &  53.02   SMH-RandForest &  47.54   SMH-Logistic \\
 69.09   BLEU &  63.74   TER &  51.12   SMH-IBk &  46.06   SimpleNets-RNN2 \\
 68.82   SimpleNets-MLP &  63.54   SimpleNets-MLP &  49.96   SimpleNets-RNN3 &  45.71   \textbf{AdaBoostClassifier} \\
 68.36   TER &  62.82   BLEU &  49.81   SimpleNets-MLP &  44.50   SMH-RandForest \\
 67.60   \textbf{GradientBoosting} &  62.72   MT-baseline &  48.31   MT-baseline &  40.94   METEOR \\
 67.53   MS-RandForest &  62.69   IIT-Meteor &  47.84   MS-IBk-b &  40.75   SimpleNets-RNN3 \\
 67.50   IIT-LM &  61.71   MS-IBk-b &  47.82   MS-RandForest &  39.85   MS-RandForest \\
 66.79   WER &  61.50   MS-IBk &  47.47   SimpleNets-RNN2 &  39.80   DeepIndiBow \\
 66.75   MS-RandForest-b &  60.38   \textbf{GradientBoosting} &  43.46   IIT-S &  39.30   IIT-Metrics \\
 65.89   DeepIndiBow &  60.12   METEOR &  42.57   DeepIndiBow &  38.27   MS-IBk \\
 65.89   DeepBow &  59.69   SMH-RandForest-b &  40.92   UoW &  38.16   MS-IBk-b \\
 65.89   MT-baseline &  59.06   WER &  39.68   Majority-class &  38.03   DeepBow \\
 65.89   Majority-class &  58.83   UoW &  38.10   MS-IBk &  37.49   MT-baseline \\
 65.72   METEOR &  51.29   SimpleNets-RNN2 &  35.58   DeepBow &  34.08   TER \\
 65.50   SimpleNets-RNN2 &  51.00   CLaC-RF &  34.88   CLaC-RF-0.5 &  34.06   CLaC-0.5 \\
 65.11   SimpleNets-RNN3 &  46.64   SimpleNets-RNN3 &  34.66   CLaC-RF-0.6 &  33.69   SimpleNets-MLP \\
 64.39   CLaC-RF-Perp &  46.30   DeepBow &  34.48   WER &  33.04   IIT-Default \\
 62.00   MS-IBk &  42.53   DeepIndiBow &  34.30   CLaC-RF-0.7 &  32.92   BLEU \\
 46.32   UoW &  42.51   Majority-class &  33.52   TER &  32.88   CLaC-0.7 \\
 &  &  33.34   METEOR &  32.20   CLaC-0.6 \\
 &  &  33.00   BLEU &  31.28   WER \\
 &  &  &  26.53   Majority-class \\
\bottomrule
\end{tabular}
\caption{Weighted F1 Score for classifiers (assign the label Good, OK or Bad)}
\label{leaderboard_f1}
\end{subtable}
\caption{QATS leaderboard. Results in \textbf{bold} are our additions to the original leaderboard. We only select the two models that rank highest during cross-validation.} \label{leaderboard}
\end{table*}

\paragraph{Pearson correlation for regressors (raw scoring)}
Combining the features does not bring a clear advantage over the elementary metrics METEOR and NBOutputSyllablesPerSent.
Indeed our best models score respectively on grammaticality, meaning preservation and simplicity: 0.33 (Lasso), 0.58 (Ridge) and 0.49 (Ridge) versus 0.39 (METEOR), 0.58 (METEOR) and 0.49 (NBOutputSyllablesPerSent).

It is surprising to us that the aggregation of multiple elementary features would score worse than the features themselves. However, we observe a strong discrepancy between the scores obtained  on the train and test set, as illustrated by TABLE~\ref{individual_features}. We also observed very large confidence intervals in terms of Pearson correlation. For instance our lasso model scores $0.33 \pm 0.17$ on the test set for grammaticality. This should observe caution when interpreting Pearson scores on QATS.

\paragraph{F1-score for classifiers (assigning labels)}
On the classification task, our models seem to score best for meaning preservation, simplicity and overall, and third for grammaticality.
This seems to confirm the importance of considering a large ensemble of elementary features including length-based metrics to evaluate simplicity.

\section{Conclusion}
Finding accurate ways to evaluate text simplification (TS) without the need for reference data is a key challenge for TS, both for exploring new approaches and for optimizing current models, in particular those relying on unsupervised, often MT-inspired models.

We explore multiple reference-less quality evaluation methods for automatic TS systems, based on data from the 2016 QATS shared task. We rely on the three key dimensions of the quality of a TS system: grammaticality, meaning preservation and simplicity.

Our results show that grammaticality and meaning preservation are best assessed using $n$-gram-based MT metrics evaluated between the output and the source sentence. In particular, METEOR and smoothed BLEU achieve the highest correlation with human judgments.
These approaches even outperform metrics that make an extensive use of external data, such as language models. This shows that a lot of useful information can be obtained from the source sentence itself.

Regarding simplicity, we observe that counting the number of characters, syllables and words provides the best results.
In other words, given the currently available metrics, the length of a sentence seems to remain the best available proxy for its simplicity.

However, given the small size of the QATS dataset and the high variance observed in our experiments, these results must be taken with a pinch of salt and will need to be confirmed on a larger dataset.
Creating a larger annotated dataset as well as averaging multiple human annotations for each pair of sentences would help reducing the variance of the experiments and confirming our findings.

In future work, we shall explore richer and more complex features extracted using syntactic and semantic analyzers, such as those used by the SAMSA metric, and paraphrase detection models.

Finally, it remains to be understood how we can optimize the trade-off between grammaticality, meaning preservation and simplicity, in order to build the best possible comprehensive TS metric in terms of correlation with human judgments.
Unsurprisingly, optimizing one of these dimensions often leads to lower results on other dimensions \cite{schwarzer2018human}.
For instance, the best way to guarantee grammaticality and meaning preservation is to leave the source sentence unchanged, thus resulting in no simplification at all. Improving TS systems will require better global TS evaluation metrics.
This is especially true when considering that TS is in fact a multiply defined task, as there are many different ways of simplifying a text, depending on the different categories of people and applications at whom TS is aimed.

\section*{Acknowledgments}
We would like to thank our anonymous reviewers for their insightful comments.

\bibliography{acl2018}
\bibliographystyle{acl_natbib}

\end{document}